\documentclass[10pt,twocolumn,letterpaper]{article}

\usepackage{cvpr}      % To produce the REVIEW version
\usepackage{graphicx}
\usepackage{amsmath}
\usepackage{amssymb}
\usepackage{booktabs}

\usepackage[pagebackref,breaklinks,colorlinks]{hyperref}

\usepackage[capitalize]{cleveref}
\crefname{section}{Sec.}{Secs.}
\Crefname{section}{Section}{Sections}
\Crefname{table}{Table}{Tables}
\crefname{table}{Tab.}{Tabs.}

\def\cvprPaperID{*****} % *** Enter the CVPR Paper ID here
\def\confName{CVPR Epic/Ego4D Workshop}
\def\confYear{2023}

\begin{document}

\title{Situated Cameras, Situated Knowledges: Towards an Egocentric Epistemology for Computer Vision}

\author{Samuel Goree\\
Indiana University\\
%Bloomington, Indiana, USA\\
%{\tt\small sgoree@iu.edu}
\and
David Crandall\\
Indiana University\\
%Bloomington, Indiana, USA\\
%{\tt\small djcran@indiana.edu}
}
\maketitle

%%%%%%%%% ABSTRACT
\begin{abstract}
   In her influential 1988 paper, \emph{Situated Knowledges}, Donna Haraway uses vision and perspective as a metaphor to discuss scientific knowledge. Today, egocentric computer vision discusses many of the same issues, except in a literal vision context. In this short position paper, we collapse that metaphor, and explore the interactions between feminist epistemology and egocentric CV as ``Egocentric Epistemology.'' Using this framework, we argue for the use of qualitative, human-centric methods as a complement to performance benchmarks, to center both the literal and metaphorical perspective of human crowd workers in CV.
\end{abstract}

\vspace{-0.2in}

%%%%%%%%% BODY TEXT
\section{Introduction}
\label{sec:intro}

In Computer Vision (CV), egocentric vision is meant very literally: processing imagery taken from a camera on a human \cite{damen2018scaling}, robot \cite{martin2021jrdb}, animal \cite{ladha2013dog} or car \cite{yao2019egocentric}. This mode of data collection creates many challenging scientific and engineering problems, particularly because unlike photography-based CV, where the camera is outside the scene and imagery is taken deliberately, egocentric data comes from a camera within the scene, captured automatically \cite{fathi2011understanding,bambach2015survey}.

In this paper, however, we  consider egocentric vision more metaphorically and think about its theoretical foundations, particularly its \emph{epistemology} --- its theory of knowledge and knowing. We believe that ongoing crises regarding bias, unfairness and injustice throughout artificial intelligence \cite{mehrabi2021survey} motivate critical epistemological investigation. In other words, we may be able to avoid problems related to bias, unfairness and injustice by developing a more robust theory of what we know about the functionality of our models and algorithms and how we know it. 

We believe there is a natural theoretical match for egocentric CV in feminist\footnote{By feminist, we refer to the philosophical tradition which emerges from feminist studies. We are not discussing sexism in computer vision.} epistemology, particularly Donna Haraway's theory of situated knowledges \cite{haraway1988situated}. This match is interesting because Haraway uses vision as a metaphor to talk about science. By collapsing this metaphor, we arrive at a natural fit for the technical reality of egocentric vision: just as egocentric vision moves the camera into the scene, an egocentric epistemology would move evaluation into the world of our participants, yielding epistemic power to them.

\begin{figure}
    \centering
    \includegraphics[width=\columnwidth]{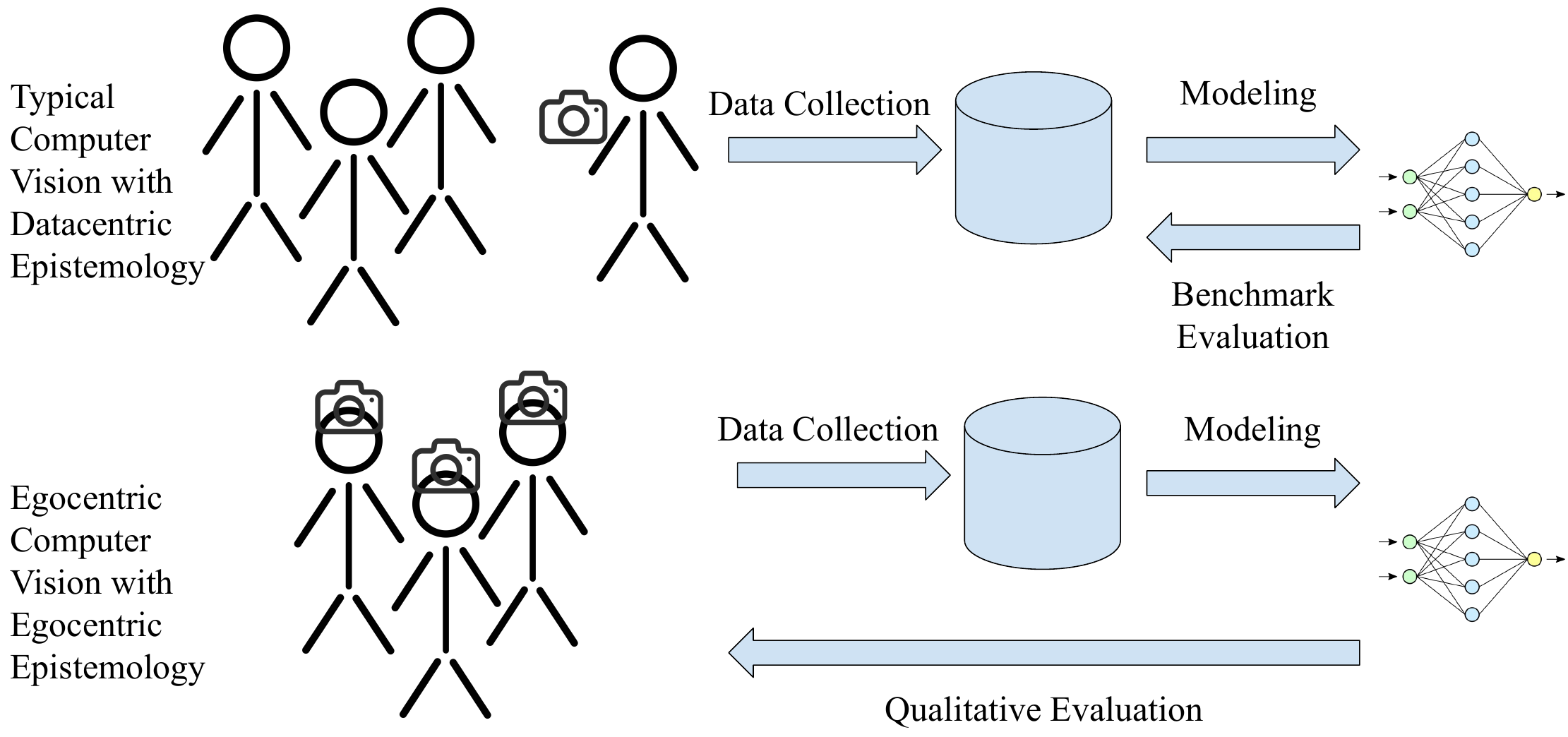}
    \caption{Just as egocentric CV places the camera in the scene, we propose placing evaluation among the research participants.}
    \label{fig:cartoon}
\end{figure}

\section{\emph{Situated Knowledges}}

In an influential 1988 paper
%, ``Situated Knowledges: The Science Question in Feminism and the Privilege of Partial Perspective,''
\cite{haraway1988situated}, Donna Haraway discusses scientific objectivity and its fraught relationship with feminism. Many of the central questions of feminist theory involve questioning scientific ``facts'' --- particularly those about women and their inferiority to men. Some feminists cite these conflicts as justification to throw out scientific inquiry itself as biased, but Haraway disagrees. She does not want a feminist critique of science to serve as ``one more excuse for [feminists] not learning any post-Newtonian physics.'' Instead, she seeks to find a feminist way of thinking which admits both real scientific knowledge as well as arguments against sexist findings. 

To reconcile these perspectives, Haraway employs vision as a metaphor. She observes that science, when it separates a ``view'' of the world from the way that it was captured, performs a ``god trick'' --- pretending that an observer's limited view can actually see everything from an omniscient god's-eye view. But all vision --- human, animal or machine --- is actually situated, limited and partial. We cannot see distant stars, bacteria or atoms as they truly are; we can only see them as they are captured by cameras, telescopes or other sensors and processed through data analysis systems to produce images designed specifically for our eyes. In other words, humans and our hybrid technological-biological vision systems are always part of the universe observing itself. Haraway argues that acknowledging the embodied and situated reality of our vision does not make the science we do on it subjective or relative, but actually makes it \emph{more objective} because we acknowledge the reality of our knowledge production. In Haraway's words:
\textit{``Infinite vision is an illusion, a god trick...We need to learn in our bodies, endowed with primate color and stereoscopic vision, how to attach the objective to our theoretical and political scanners...Objectivity turns out to be about particular and specific embodiment and definitely not about the false vision promising transcendence of all limits and responsibility. The moral is simple: only partial perspective promises objective vision. All Western cultural narratives about objectivity are allegories of the ideologies governing the relations of what we call mind and body, distance and responsibility. Feminist objectivity is about limited location and situated knowledge, not about transcendence and splitting of subject and object. It allows us to become answerable for what we learn how to see.''}
%\end{quote}

Haraway's position is not an attack on scientists, who  typically acknowledge the limitations of their instruments and  methods. Rather it is a critique of industry, government and the public, who perform god tricks when they treat the findings of scientists as completely true, detached from the limitations of their research methods.

\section{Two God Tricks in CV}

Traditional CV engages in two different kinds of god tricks. First, it treats sets of images as objective recordings of reality, detached from the cameras and photographers who take them and the researchers who assemble them. Second, it treats its knowledge about the performance of models and algorithms as objective truth, separate from the data and methods which allow us to evaluate them. 

The combination of these two practices has had horrible consequences with respect to bias and injustice in CV. Algorithms which are purported to be objective and neutral reflect a dominant American cultural perspective which exists both in their training data, as well as in the academic-industrial research system which produces the models and algorithms \cite{denton2021genealogy}. Many CV systems have substantial limitations: they are only ever approximately correct, only have limited knowledge of the world and faithfully reproduce the biases, both good and bad, of their training data \cite{buolamwini2018gender}. 

While researchers often identify these limitations, the science enterprise and broader public have been conditioned by futureological interpretations of science fiction to treat these systems as simultaneously human-like \cite{hermann2023artificial} and omniscient, indulging the dreams of the military intelligence-industrial complex \cite{suchman2022imaginaries}. Unfortunately, in pursuit of research funding and employment, CV researchers are not incentivized to acknowledge
the limitations of their work
%fuel those fantasies, exaggerating the success and usefulness of modeling and leaving limitations understated or unstudied 
\cite{goree2022attention}. The problem is not the existence of limitations, but the way the research system performs a god trick and transforms algorithmic tools which provide situated, uncertain knowledge about the world into arbiters of objective truth. For example, a system designed to recognize faces which systematically misidentifies nonwhite and nonmale faces \cite{buolamwini2018gender} is not inherently harmful. Rather it could become harmful when its high accuracy is equated with objectively good performance, justifying its use in production systems.

\subsection{Egocentric Vision Avoids the First God Trick}

Like Haraway believes there is a way to criticize bias in science without rejecting scientific knowledge, we believe there is a way to criticize bias in CV without rejecting knowledge about technical performance. Egocentric CV is naturally suited towards this reconciliation because it avoids the first god trick: egocentric images are messy: the cameras shake and scenes are often partially obscured. The images usually contain hands \cite{bambach2015lending} and sometimes include other observers who have their own cameras \cite{fan2017identifying}.  In contrast, images taken by human photographers usually come from outside the scene they depict. They are well-framed, with objects un-occluded. The photographer can control exposure time and focal length to best represent the scene \cite{hertzmann2022choices}. Counter-intuitively,  egocentric images are often more objective, less authored, views of a scene because they avoid the god trick of the photograph. They depict the world more like it appears to a particular human, not the way a photographer believes it should be depicted.

But egocentric vision still takes part in the second god trick. We treat evaluations using quantitative metrics on benchmark datasets as true, a view from above which provides objective evaluation of the relative strengths and weaknesses of our models. However, our performance metrics are more like a photographers' camera: they are designed by CV researchers, sometimes the same researchers designing the models under evaluation. Those researchers make numerous decisions regarding the collection and curation of the data, and define what good performance means, with their own external goals and applications in mind. This approach is not objective, but that is not a bad thing! Just as there is no digital image without a camera or sensor to capture it, there is no problem statement or dataset without a human author and underlying motive. To some extent that is good --- our motives for proposing vision problems ground them in reality and make their solutions valuable. But the task definition overwrites the normative perspective of the person wearing the camera with that of the researchers.

\subsection{Avoiding the Second God Trick}

In the same way that egocentric vision avoids the first god trick by situating cameras on human bodies, it can avoid the second god trick by situating the evaluation process in human experience. Human evaluation has precedent in an earlier period of CV research before the rise of large datasets, when research papers would demonstrate effectiveness by showing sample image results
\cite{goree2022attention}. But rather than leave evaluation to the judgments of conference reviewers, we propose collecting evaluations from the same people who provide our training data. 

To envision what an  egocentric epistemology might look like, we can look to the social sciences. Methods grounded in this tradition are typically qualitative and ethnographic, studying humans by entering their social worlds. These methods are not new to computing: feminist and poststructuralist approaches are being explored in HCI \cite{bardzell2011towards} and in data science \cite{d2020data}. However, we cannot follow HCI and center the user in CV because  our algorithms might be used in many different user-facing applications, or in applications without human users at all. So rather than center the user, we advocate for centering the crowd workers who collect our data.

To envision what this approach would look like, we  turn to secondary literature which operationalizes Haraway's theory. Bhavnani \cite{bhavnani1993tracing} proposes three criteria:
\begin{enumerate}
    \item Reinscription: Does the research method portray the participants as passive and powerless, or does it recast them as active agents?
    \item Micropolitics: Does the research engage with the political relationships between researcher and participant?
    \item Difference: Does the research engage with differences in perspective between participants?
\end{enumerate} 
\noindent
We can apply these principles to CV evaluation. For example, in the context of activity recognition on an Epic Kitchens-style dataset \cite{damen2018scaling} we could evaluate several activity detectors by returning to the initial participants captured on video, demonstrate various activity recognition models, discuss how they interpret the participants' behaviors and select a ``state of the art.'' This fits Bhavnani's criteria:

\begin{enumerate}
    \item Reinscription: Treat crowd workers as active agents in the research work, rather than passive data producers.
    \item Micropolitics: Acknowledge the power differential between researchers and crowd workers, and yield governance over the modeling process to those workers.
    \item Difference: When participants disagree with each other or with researchers about evaluation, it can teach us about the nuances of the task, and is not just noise.
\end{enumerate}

Of course, going back to the same initial crowd workers every time we propose a new model architecture is impractical. Such an approach does not scale with the fast pace of vision research. CV researchers are also not social scientists, and should not have to retrain to carry out our own research. But returning to participants once or twice after collecting data can be a valuable supervisory signal, and counteract the unequal power dynamics involved in data collection in CV \cite{paullada2021data,denton2021genealogy}. We encourage collaborating with scholars who have qualitative research training when working with human subjects to remain systematic and impartial.

While this kind of research is qualitative, it can be highly rigorous. Sometimes participants might say, ``well that looks wonderful!'' or ``no, I don’t think computers should do that.'' While these are legitimate answers, they are ultimately unhelpful. Forcing participants to choose between two alternatives is also unhelpful: we want to know the problems with those alternatives and why they matter. We will often find by talking to people that the problem we are trying to address is not a helpful problem to solve, or that a less technically demanding method is actually more helpful. These are productive research findings! While they do not generate scientific advances in our modeling work, they help us avoid wasting time, effort and GPU-hours on unhelpful problems. Counter-intuitively, by taking part in seemingly inefficient, human-centric research, we can actually make our technical work more impactful, and teach our students to think critically about modeling tasks.

\section{An Example:  Aesthetic Quality Assessment}

Driven by these concerns, we have recently investigated taking an egocentric epistemology in the evaluation of a non-egocentric CV problem: aesthetic quality assessment (AQA), or the task of determining whether a photograph is high or low aesthetic quality. Over the past  decades, benchmarked performance on the Analysis of Visual Aesthetics (AVA) dataset \cite{murray2012ava} has increased steadily, but it remains uncertain whether these models detect aesthetic quality or  confounding stylistic factors \cite{goree2021does}. To ground evaluation for this problem in the aesthetic experience of human participants, we designed a smartphone-based camera interface without a shutter button. Photos are taken when the response of an AQA model exceeds an adaptive threshold.

In ongoing research, we are conducting a series of semi-structured user testing sessions with our interface and four candidate models, inspired by feature and model architectures used over the history of AQA. Participants explore the similarities and differences between models while walking around a public space, seeing which models capture pictures of which objects and scenes. At the end, we review the photos taken by each model, and ask participants to evaluate the models' strengths and weaknesses. These sessions have been highly informative, revealing the biases of each model towards specific patterns, objects or compositions. Further, these sessions have allowed users to give critical feedback on the problem formulation and its assumptions. These aspects are often left unquestioned except by coauthors and peer reviewers. Counter-intuitively, by engaging in human-centric evaluation, we end up with an approach to evaluation which is less limited by the views of CV researchers.

\section{Conclusion}

In this position paper, we have discussed Donna Haraway's theory of situated objectivity. We have identified two ways that CV uses ``god tricks'' to pretend to see the world from nowhere. Egocentric cameras avoid one  but fall victim to the other, so we propose an  egocentric epistemology which returns evaluation to the site of data collection.

Adopting human-centric methods in egocentric vision is only a first step towards a foundation for CV which preempts problems related to bias and injustice. No amount of talking to human participants matters unless we believe and act on what they have to say. Additionally, if we only engage with certain kinds of participants (such as the WEIRD participants of psychology \cite{henrich2010most}), our knowledge will be limited to the perspectives of the people we approach. Giving our participants evaluation power over the modeling process is a step towards counteracting larger problems related to power and surveillance in CV \cite{crawford2021atlas}, but developing a model using ethical research methods does not guarantee that applications of that model are ethical. Still, we encourage further discussion of power dynamics in CV and ways that we can adjust our research methods to preempt future issues related to bias and injustice in our discipline.

%%%%%%%%% REFERENCES
{\small
\bibliographystyle{abbrv}
\bibliography{egbib}

\begin{thebibliography}{10}

\bibitem{bambach2015survey}
S.~Bambach.
\newblock A survey on recent advances of computer vision algorithms for
  egocentric video.
\newblock {\em arXiv}, 2015.

\bibitem{bambach2015lending}
S.~Bambach, S.~Lee, D.~J. Crandall, and C.~Yu.
\newblock Lending a hand: Detecting hands and recognizing activities in complex
  egocentric interactions.
\newblock In {\em ICCV}, pages 1949--1957, 2015.

\bibitem{bardzell2011towards}
S.~Bardzell and J.~Bardzell.
\newblock Towards a feminist hci methodology: social science, feminism, and
  hci.
\newblock In {\em CHI}, pages 675--684, 2011.

\bibitem{bhavnani1993tracing}
K.-K. Bhavnani.
\newblock Tracing the contours: Feminist research and feminist objectivity.
\newblock In {\em Women's Studies International Forum}, volume~16, pages
  95--104. Elsevier, 1993.

\bibitem{buolamwini2018gender}
J.~Buolamwini and T.~Gebru.
\newblock Gender shades: Intersectional accuracy disparities in commercial
  gender classification.
\newblock In {\em FAccT}, pages 77--91. PMLR, 2018.

\bibitem{crawford2021atlas}
K.~Crawford.
\newblock {\em The atlas of AI: Power, politics, and the planetary costs of
  artificial intelligence}.
\newblock Yale University Press, 2021.

\bibitem{damen2018scaling}
D.~Damen, H.~Doughty, G.~M. Farinella, S.~Fidler, A.~Furnari, E.~Kazakos,
  D.~Moltisanti, J.~Munro, T.~Perrett, W.~Price, et~al.
\newblock Scaling egocentric vision: The epic-kitchens dataset.
\newblock In {\em ECCV}, pages 720--736, 2018.

\bibitem{denton2021genealogy}
E.~Denton, A.~Hanna, R.~Amironesei, A.~Smart, and H.~Nicole.
\newblock On the genealogy of machine learning datasets: A critical history of
  imagenet.
\newblock {\em Big Data \& Society}, 8(2), 2021.

\bibitem{d2020data}
C.~D'ignazio and L.~F. Klein.
\newblock {\em Data feminism}.
\newblock MIT press, 2020.

\bibitem{fan2017identifying}
C.~Fan, J.~Lee, M.~Xu, K.~Kumar~Singh, Y.~Jae~Lee, D.~J. Crandall, and M.~S.
  Ryoo.
\newblock Identifying first-person camera wearers in third-person videos.
\newblock In {\em CVPR}, pages 5125--5133, 2017.

\bibitem{fathi2011understanding}
A.~Fathi, A.~Farhadi, and J.~M. Rehg.
\newblock Understanding egocentric activities.
\newblock In {\em ICCV}, pages 407--414. IEEE, 2011.

\bibitem{goree2021does}
S.~Goree.
\newblock What does it take to cross the aesthetic gap? the development of
  image aesthetic quality assessment in computer vision.
\newblock In {\em ICCC}, pages 11--15, 2021.

\bibitem{goree2022attention}
S.~Goree, G.~Appleby, D.~Crandall, and N.~Su.
\newblock Attention is all they need: Exploring the media archaeology of the
  computer vision research paper.
\newblock {\em arXiv}, 2022.

\bibitem{haraway1988situated}
D.~Haraway.
\newblock Situated knowledges: The science question in feminism and the
  privilege of partial perspective.
\newblock {\em Feminist studies}, 14(3):575--599, 1988.

\bibitem{henrich2010most}
J.~Henrich, S.~J. Heine, and A.~Norenzayan.
\newblock Most people are not weird.
\newblock {\em Nature}, 466(7302):29--29, 2010.

\bibitem{hermann2023artificial}
I.~Hermann.
\newblock Artificial intelligence in fiction: between narratives and metaphors.
\newblock {\em AI \& society}, 38(1):319--329, 2023.

\bibitem{hertzmann2022choices}
A.~Hertzmann.
\newblock The choices hidden in photography.
\newblock {\em Journal of Vision}, 22(11):10--10, 2022.

\bibitem{ladha2013dog}
C.~Ladha, N.~Hammerla, E.~Hughes, P.~Olivier, and T.~Ploetz.
\newblock Dog's life: wearable activity recognition for dogs.
\newblock In {\em UBICOMP}, pages 415--418, 2013.

\bibitem{martin2021jrdb}
R.~Martin-Martin, M.~Patel, H.~Rezatofighi, A.~Shenoi, J.~Gwak, E.~Frankel,
  A.~Sadeghian, and S.~Savarese.
\newblock Jrdb: A dataset and benchmark of egocentric robot visual perception
  of humans in built environments.
\newblock {\em PAMI}, 2021.

\bibitem{mehrabi2021survey}
N.~Mehrabi, F.~Morstatter, N.~Saxena, K.~Lerman, and A.~Galstyan.
\newblock A survey on bias and fairness in machine learning.
\newblock {\em CSUR}, 54(6):1--35, 2021.

\bibitem{murray2012ava}
N.~Murray, L.~Marchesotti, and F.~Perronnin.
\newblock Ava: A large-scale database for aesthetic visual analysis.
\newblock In {\em CVPR}, pages 2408--2415. IEEE, 2012.

\bibitem{paullada2021data}
A.~Paullada, I.~D. Raji, E.~M. Bender, E.~Denton, and A.~Hanna.
\newblock Data and its (dis) contents: A survey of dataset development and use
  in machine learning research.
\newblock {\em Patterns}, 2(11):100336, 2021.

\bibitem{suchman2022imaginaries}
L.~Suchman.
\newblock Imaginaries of omniscience: Automating intelligence in the us
  department of defense.
\newblock {\em Social Studies of Science}, 2022.

\bibitem{yao2019egocentric}
Y.~Yao, M.~Xu, C.~Choi, D.~J. Crandall, E.~M. Atkins, and B.~Dariush.
\newblock Egocentric vision-based future vehicle localization for intelligent
  driving assistance systems.
\newblock In {\em ICRA}, pages 9711--9717. IEEE, 2019.

\end{thebibliography}
}

\end{document}